\documentclass[conference]{IEEEtran}
\IEEEoverridecommandlockouts

\usepackage{cite}
\usepackage{amsmath,amssymb,amsfonts}
\usepackage{algorithmic}
\usepackage{textcomp}
\usepackage{booktabs}
\usepackage{amsfonts}
\usepackage{amsmath}
\usepackage{amssymb}
\usepackage{amsthm}
\usepackage{microtype}
\usepackage{xcolor}
\usepackage{graphicx}
\usepackage{multirow}
\usepackage{lipsum}
\usepackage{wrapfig}
\usepackage{float}
\usepackage{orcidlink}

\def\BibTeX{{\rm B\kern-.05em{\sc i\kern-.025em b}\kern-.08em
    T\kern-.1667em\lower.7ex\hbox{E}\kern-.125emX}}
\begin{document}

\title{STMemoLLM: A Spatial-Temporal Memory-Augmented LLM for Robust Traffic Forecasting}


\author{
\IEEEauthorblockN{
Jian Zhou\textsuperscript{1,*} \orcidlink{0009-0005-0773-3447},
Binqing Wu\textsuperscript{2,*}\orcidlink{0000-0001-8276-0801},
Zongjiang Shang\textsuperscript{3} \orcidlink{0009-0001-8938-7437},
Wanghui Qiu\textsuperscript{2} \orcidlink{0009-0006-1896-6913},
Dongliang Cui\textsuperscript{1} \orcidlink{0000-0002-8306-0513}, \\
Chenxi Jin\textsuperscript{1}\orcidlink{0009-0007-1133-454X},
Yue Yu\textsuperscript{1} \orcidlink{0009-0005-8655-6265},
Xinyi Fu\textsuperscript{2}\orcidlink{0000-0002-5260-3992},
~and Ling Chen\textsuperscript{1,\dag} \orcidlink{0000-0003-1934-5992}
}
\thanks{*Equal Contribution}
\thanks{\dag Corresponding Author}
\IEEEauthorblockA{
\textsuperscript{1}Zhejiang University, Hangzhou, China.
\textsuperscript{2}Ant Group, Hangzhou, China. 
\textsuperscript{3}Hunan University, Changsha, China.
}
\IEEEauthorblockA{
22521068@zju.edu.cn, wubinqing.wbq@antgroup.com, shangzongjiang@hnu.edu.cn, \\
qiuwanghui.qwh@antgroup.com, 12321057@zju.edu.cn, 22521163@zju.edu.cn,\\
22421234@zju.edu.cn, fxy122992@antgroup.com, lingchen@cs.zju.edu.cn}
}

\maketitle

\begin{abstract}
Traffic forecasting is essential for intelligent transportation systems.
However, traditional deep-learning models often rely heavily on high-quality observations. Recent LLM-based approaches incorporate pretrained knowledge to improve forecasting, but they remain sensitive to input perturbations and struggle to preserve long-term patterns.
To address this, we propose STMemoLLM, a spatial–temporal memory-augmented LLM for robust traffic forecasting. STMemoLLM introduces memory prototypes to compress inputs into compact summary tokens while extracting reliable dynamics from residual signals, achieving a balance between stability and variability.
Specifically, a memory-enhanced tokenizer encodes prototypes and sensor characteristics into attention biases to guide adaptive compression of correlated sensors into semantic summary tokens. A memory-calibrated router further evaluates contextual deviation, compression deficiency, and prototype alignment to identify reliable residual signals, which complement the summaries while suppressing noise.
Extensive experiments on eight real-world datasets show that STMemoLLM achieves state-of-the-art performance. Additional experiments examine point masking, additive Gaussian noise, structured missingness, and limited training data, demonstrating strong robustness. 
Code is available at https://anonymous.4open.science/r/STMemoLLM-70BF.
\end{abstract}


\section{Introduction}
Traffic forecasting is a cornerstone of modern intelligent transportation systems, supporting critical applications including route planning, congestion mitigation, and urban infrastructure management \cite{stgnn_survey,traff_survey_2,traff_survey}. 
Although substantial progress has been made, evolving from statistical approaches \cite{arima,var} to deep learning methods \cite{sun2027dualefficient,jin2027robusttraffic}, significant challenges still remain in effectively modeling the complex spatial and temporal patterns inherent in traffic systems.

Over the past five years, spatial-temporal graph neural networks (STGNNs) have established the state-of-the-art (SOTA) performance in traffic forecasting by modeling spatial patterns with graph neural networks and temporal patterns with sequential models \cite{STDN,megacrn}. In parallel, Transformer-based approaches have demonstrated strong potential, as self-attention enables a global receptive field that captures long-range spatial and temporal patterns \cite{an2026dambast,pdformer}. Nevertheless, these methods rely heavily on high-quality data to learn such patterns, which makes them susceptible to data sparsity (e.g., incomplete sensor recordings) and noise (e.g., measurement inaccuracies). In practice, traffic observations are often incomplete or noisy due to missing recordings, measurement errors, and sensor failures. Robust forecasting therefore requires models to capture stable traffic patterns while limiting the influence of unreliable observations.

Recent studies have explored pretrained large language models (LLMs), which leverage rich prior knowledge to learn traffic patterns \cite{foundation_survey,foundation_survey_1}. By reformulating time series as language-like sequences, these methods map numerical observations into a semantically rich representation space, allowing patterns that are difficult to detect in low-quality data to emerge through the pretrained temporal reasoning capabilities of models \cite{timellm,onefitsall}. Some LLM-based methods further strengthen domain awareness. Some incorporate richer geographical context (e.g., points of interest (POIs), coordinates, and addresses) via instruction or prompt engineering \cite{fstllm,he2025geolocation,urbangpt}. Others fine-tune LLMs by employing distance-guided attention masks or by directly encoding network connectivity into the positional embeddings \cite{liu2025st,Std-plm,stllm}. Nevertheless, these methods rely on pretrained reasoning over real-time inputs. While this enables them to capture dynamic patterns, they often fail to retain stable patterns persistent in the dataset. Consequently, the model may be sensitive to sample-specific fluctuations.

Besides leveraging pretrained priors, a complementary line of research explicitly memorizes stable traffic patterns persistent in the dataset through learnable memory prototypes \cite{megacrn,gao2025how}. Each prototype represents a characteristic traffic pattern shared across training samples (e.g., high/low flow and volatile/stable conditions). More recently, some studies have further integrated memory banks into LLM-based traffic forecasting, combining pretrained priors with explicit traffic pattern memories to enhance hidden representations within LLMs \cite{jeon2025st}. However, these memory mechanisms primarily determine prototype relevance through input-dependent similarity, with limited explicit consideration of persistent sensor-specific preferences. Prototype matching may therefore be sensitive to transient or corrupted observations. Moreover, similarity to stored patterns does not directly establish the reliability of variations beyond them, which may reflect either meaningful traffic changes or stochastic noise.

These limitations motivate us to use memory prototypes to guide the construction of LLM inputs that preserve recurring traffic patterns while retaining reliable dynamics from current observations. Realizing this goal presents two challenges:
(1) \textbf{Sensor-specific prototype preferences}. Many methods retrieve prototypes based on input-dependent Query-Key similarities \cite{megacrn,jeon2025st}, neglecting the fact that different sensors have distinct inherent preferences for prototypes. For example, a “morning rush” pattern may be retrieved for all sensors, but its predictive usefulness depends heavily on sensors' spatial context, e.g., producing sharp peaks in commercial districts and subtle fluctuations in residential areas. Without explicit sensor-specific preferences, the LLM cannot prioritize prototypes that best match the unique characteristics of each sensor, making retrieval overly susceptible to fluctuations.
(2) \textbf{Reliable dynamics beyond prototypes}. 
Prototype-based methods usually act as low-pass filters, distilling stable patterns while treating deviations as spurious residuals. Despite enhancing robustness, the static prototype space creates an information bottleneck that overlooks informative dynamics embedded in residuals. However, these residuals are often noise-dominated, making it challenging to select reliable dynamics from them.

To this end, we propose STMemoLLM, a spatial-temporal memory-augmented LLM to capture stable and dynamic traffic patterns for robust traffic forecasting.
Specifically, guided by memory prototypes, STMemoLLM compresses correlated sensors into summary tokens and filters reliable dynamic information from noise to supplement details, thereby striking a balance between stability and variability.
The main contributions are as follows:
\begin{itemize}
    \item We introduce a memory-enhanced tokenizer that encodes memory prototypes and sensor characteristics into attention biases, enabling the model to prioritize correlated sensors and prototypes. Guided by these biases, sensors are then compressed into a small set of semantically rich summary tokens, which can extract stable patterns and enhance computational efficiency.
    \item We introduce a memory-calibrated router that identifies reliable dynamics from residuals by evaluating contextual deviation, compression deficiency, and prototype alignment. By filtering such dynamics, this router can complement the summary tokens with additional details while maintaining resilience to stochastic noise.
    \item Extensive experimental results on eight real-world datasets demonstrate that STMemoLLM achieves the SOTA performance, outperforming the strongest baseline STD-PLM by 1.92\% in MAE on average. Moreover, STMemoLLM demonstrates consistent robustness across with average relative MAE reductions of 1.48\% under corruption, 12.46\% under structured missingness, and 7.70\% under few-shot settings, respectively.
\end{itemize}

\begin{figure*}
    \centering
    \includegraphics[width=\linewidth]{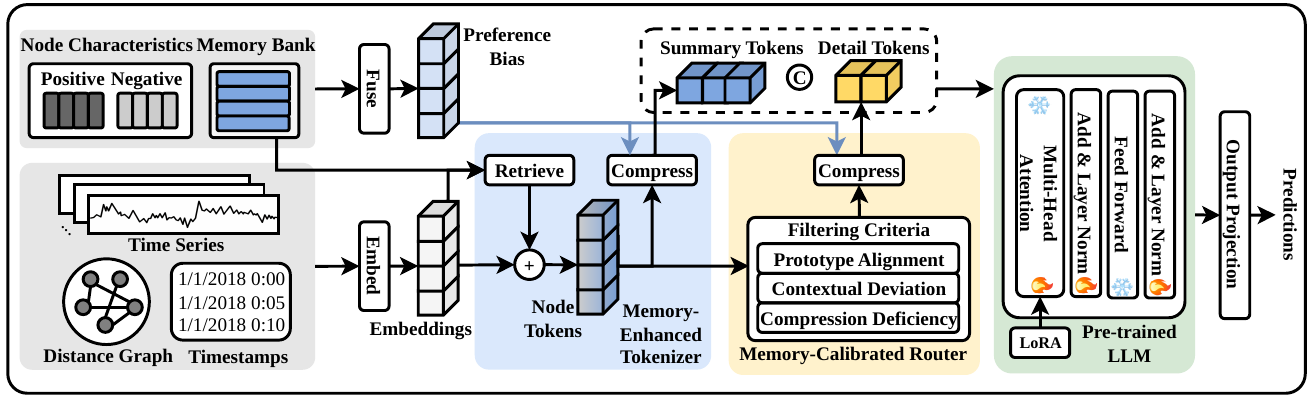}
    \caption{Overview of STMemoLLM}
    \label{fig:framework}
\end{figure*}

\section{Related Work}
Deep learning has become the dominant paradigm for traffic forecasting, overcoming the linearity assumptions of classical methods \cite{traff_survey}.
These methods have evolved from adopting Recurrent Neural Networks, Convolutional Neural Networks, and simple attention mechanisms to STGNNs \cite{luo2023stg4traffic,stgnn_survey}.
STGNNs mainly adopt a hybrid architecture: GNNs capture spatial patterns from pre-defined graphs (based on geographic proximity or road connectivity \cite{stgcn,stmgcn}) or adaptive graphs (learned from node embeddings or time similarity \cite{dstcgcn,trendgcn}), while sequential models capture temporal dynamics \cite{agcrn,megacrn}. By jointly modeling spatial and temporal patterns, STGNNs can capture topological interactions and traffic dynamics for accurate traffic forecasting, with recent efforts further enriching this line through self-supervised representation learning and higher-order expansion structures \cite{yu2025clear,qin2025taylors}.
In addition to STGNNs, Transformers have achieved competitive performance, benefiting from their strong capability in modeling long-range dependencies. They apply attention along the temporal dimension \cite{itransformer,duet}, while some variants further extend it to the spatial dimension to jointly capture spatial and temporal patterns \cite{pdformer,STAEformer}.
Nevertheless, these deep learning methods rely heavily on high-quality data, which is often difficult to guarantee in real-world traffic systems owing to missing values, noise, and sensor failures.
 
To address this limitation, recent studies have explored LLMs (e.g., GPT \cite{achiam2023gpt} and DeepSeek \cite{liu2024deepseek})
to leverage their prior knowledge to support stability \cite{foundation_survey,foundation_survey_1}. These methods typically reformulate time series by directly tokenizing numerical values \cite{onefitsall,autotimes} or aligning them with textual representations \cite{timellm,shang2026multi}.
In traffic-related scenarios, several studies have demonstrated promising performance \cite{traff_survey}. For example, UrbanGPT \cite{urbangpt} aligns traffic data with geographical context (e.g., points of interest and addresses) through instruction tuning, enabling strong zero-shot urban prediction. FSTLLM \cite{fstllm} focuses on extremely data-scarce settings and leverages geographical prompts to enhance few-shot performance. STLLM+ \cite{liu2025st} explicitly incorporates road network structures as attention masks within LLM layers and fine-tunes the model for better adaptation. STD-PLM \cite{Std-plm} encodes road network structures into positional embeddings, facilitating the inductive learning of road topology. LagLLM \cite{lagllm} tokenizes series with lag features and unifies data-driven dynamics modeling and knowledge-driven semantic reasoning. Despite the achievement, these approaches mainly tokenize inputs without explicit pattern memories, leaving the encoded representations still sensitive to sample-specific fluctuations.

Besides leveraging pretrained priors, some works explicitly memorize recurring patterns to stabilize forecasting.
MegaCRN \cite{megacrn} augments graph convolutions with a node memory that retrieves spatio-temporal patterns on the fly.
MPM \cite{wang2025mpm} maintains multiple pattern memories for short-term forecasting, and heterogeneous experts with advantageous pattern memory \cite{yao2025patternexperts} retrieve traffic patterns to coordinate expert specialization.
M$^2$Net \cite{zhou2027m2net} further separates global shared prototypes from node-specific residuals via prototype residual learning, which closely resembles our decomposition of stable summary tokens and dynamic detail tokens.
Li-Net \cite{gao2026linet} compresses abundant channels into compact tokens via top-$k$ sparse attention. These methods memorize recurring regularities, yet the reliable dynamics beyond the prototypes are either preserved together with noise or oversmoothed away. More recently, a few LLM-based methods also integrate memory with pretrained priors. For example, ST-LINK \cite{jeon2025st} encodes spatial dependencies into rotary position embeddings and retrieves historical patterns via a memory-augmented feed-forward network. However, the retrieved patterns reinforce all representations indiscriminately, without distinguishing reliable dynamics from noise.

Therefore, we propose STMemoLLM, which leverages memory prototypes to compress inputs into stable tokens while filtering reliable dynamics from residual signals, thereby balancing stability and variability.

\section{Methodology}
\subsection{Task Formulation} 
Traffic forecasting aims to predict future traffic data (e.g., traffic flow, speed, or demand) of a road network given historical observations \cite{traff_survey,traff_survey_2}. Let $\boldsymbol{G} = (\boldsymbol{V}, \boldsymbol{E})$ denote a road network, where $\boldsymbol{V}$ is the set of nodes (e.g., sensors or road segments) and $\boldsymbol{E}$ is the set of edges, representing the connectivity between them. Given graph-structured traffic data $\boldsymbol{X}^{1:T} \in \mathbb{R}^{N \times T}$, where $N$ is the number of nodes and $T$ is the number of historical time steps, the task is to learn a mapping function $\mathcal{F}$ that predicts future traffic states for the next $P$ time steps:
\begin{equation}
    \hat{\boldsymbol{X}}^{T+1:T+P} = \mathcal{F}(\boldsymbol{X}^{1:T}, \boldsymbol{G};\boldsymbol{\Theta}),
\end{equation}
where $\hat{\boldsymbol{X}}^{T+1:T+P}$ denotes the predicted results. $\boldsymbol{\Theta}$ denotes the learnable parameters.

\subsection{Overview}
Fig. \ref{fig:framework} illustrates the overview of STMemoLLM. 
To jointly capture stable patterns and dynamic details, the input tokens to the LLM backbone consist of summary tokens and detail tokens.
Summary tokens are produced by a \textbf{memory-enhanced tokenizer}.
Specifically, we first fuse memory prototypes and node characteristics to obtain a preference bias in attention, which encourages grouping similar sensors. Guided by this bias, the tokenizer compresses the inputs into a compact set of summary tokens via attention, preserving dominant traffic patterns while reducing redundancy.
Detail tokens are derived from reliable dynamics filtered by a \textbf{memory-calibrated router}. Specifically, the router first evaluates the compression residuals of node tokens with (i) contextual deviation (difference of the compression residual against neighbor sensors), (ii) compression deficiency (information loss during summarization), and (iii) prototype alignment (margin between the two most similar memorized patterns) to filter reliable dynamics. Then, the router compresses the selected node tokens into detail tokens via attention, supplementing summary tokens with fine-grained information while suppressing anomalies. Finally, the integrated token sequence, preceded by two time tokens that summarize the global temporal evolution, is processed by a fine-tuned LLM, and its last hidden states are distributed back to nodes via cross-attention decoding to yield predictions. 
For notational clarity, all node-wise computations are performed in a canonical topological order of the network, and the final predictions are mapped back to the original node order.

\subsection{Memory-Enhanced Tokenizer}

\paragraph{Memory Bank}
Following established literature \cite{megacrn,gao2025how}, we construct a memory bank composed of learnable parameters.
Formally, a memory bank consists of $K$ learnable prototype vectors $\boldsymbol{M} = \{\boldsymbol{m}_1, \boldsymbol{m}_2, \dots, \boldsymbol{m}_K\}$, each serving as a $d_{\text{m}}$-dimensional template for a specific traffic state. These prototypes are optimized via backpropagation during the training phase and remain fixed during inference.

\paragraph{Node Tokens}
We first embed the inputs. The inputs comprise historical observations, a road-network adjacency graph, and timestamps. 
Observations are projected into a latent space via an MLP. The adjacency graph is processed via Laplacian eigendecomposition; the top-$k$ eigenvectors of the symmetrically normalized Laplacian are used as spectral features and linearly projected to obtain spatial embeddings $\boldsymbol{e}_{\text{s}}$. Timestamps are encoded into temporal embeddings $\boldsymbol{e}_{\text{t}}$, consisting of a 5-minute-resolution time-of-day embedding and a day-of-week embedding. The input node features are then embedded as:
\begin{equation}
\boldsymbol{H}_{\text{in}} = \operatorname{LayerNorm}(\operatorname{MLP}(\boldsymbol{X})+ \operatorname{MLP}(\boldsymbol{e}_{\text{t}}||\boldsymbol{e}_{\text{s}})),
\end{equation}
where $\boldsymbol{H}_{\text{in}} \in \mathbb{R}^{N \times d}$ denotes the node features, $\boldsymbol{X} \in \mathbb{R}^{N \times T}$ denotes the historical observations, and $\|$ denotes concatenation. We then compute attention scores between the node features and memory prototypes to retrieve relevant memory:
\begin{equation}
    \boldsymbol{a} = \operatorname{SoftMax}\!\left(\cos(\boldsymbol{H}_{\text{in}}\boldsymbol{W}_{\text{a}},\,\boldsymbol{M})/\tau\right),
\label{eq:memattn}
\end{equation}
where $\boldsymbol{W}_{\text{a}} \in \mathbb{R}^{d \times d_{\text{m}}}$ is a learnable projection matrix, $\cos(\cdot,\cdot)$ computes the row-wise cosine similarity between the projected node features and the prototypes, and $\tau$ is a temperature shared with the memory regularization loss.
$\boldsymbol{a} \in \mathbb{R}^{N \times K}$ represents the attention scores of each node over the $K$ memory prototypes.
The retrieved memory is computed as a weighted combination of the prototypes. To adaptively fuse memories with node features, we employ a gating mechanism:
\begin{equation}
    \boldsymbol{r}  =(\boldsymbol{a}\boldsymbol{M})\boldsymbol{W}_{\text{v}}, \quad \boldsymbol{g} = \sigma((\boldsymbol{H}_{\text{in}} \|\boldsymbol{r} )\boldsymbol{W}_{\text{g}}), \quad
    \boldsymbol{H}_{\text{re}} = \boldsymbol{g} \odot \boldsymbol{r} ,
\label{eq:gate}
\end{equation}
where $\boldsymbol{W}_{\text{v}} \in \mathbb{R}^{d_{\text{m}} \times d}$ and $\boldsymbol{W}_{\text{g}}\in \mathbb{R}^{2d \times d}$ are learnable projection matrices. $\boldsymbol{r} \in \mathbb{R}^{N \times d}$ denotes the memory retrieved for each node. $\sigma(\cdot)$ denotes the sigmoid function, and $\odot$ denotes element-wise multiplication.
The gating mechanism dynamically controls the contribution of retrieved memory conditioned on the node features, yielding $\boldsymbol{H}_{\text{re}} \in \mathbb{R}^{N \times d}$.
Finally, we incorporate a residual connection to obtain the node tokens:
\begin{equation}
    \boldsymbol{H}_{\text{node}} = \operatorname{LayerNorm}(\boldsymbol{H}_{\text{in}}+\boldsymbol{H}_{\text{re}}).
\label{eq:resnode}
\end{equation}

\paragraph{Preference Bias}
To model the inherent preferences of nodes toward memory prototypes, we introduce a dual-path mechanism that captures positive and negative inclinations. We introduce two learnable association matrices, $\boldsymbol{E}_{\text{pos}}, \boldsymbol{E}_{\text{neg}} \in \mathbb{R}^{N \times K}$, which assign node-specific weights over the $K$ prototypes and induce prototype-conditioned node descriptors $\boldsymbol{E}_{\text{pos}}\boldsymbol{M}, \boldsymbol{E}_{\text{neg}}\boldsymbol{M} \in \mathbb{R}^{N \times d_{\text{m}}}$.

Beyond individual preferences, we further account for spatial interference among nodes. To capture the persistent reinforcement or suppression effects under specific traffic states, we model such interference by interacting the dual-path descriptors across the network. Specifically, we introduce two interference matrices:
\begin{equation}
\begin{aligned}
    \boldsymbol{G}_{\text{pos}} & = \operatorname{SoftMax}(\operatorname{ReLU}(\boldsymbol{E}_{\text{pos}}\boldsymbol{M}(\boldsymbol{E}_{\text{neg}}\boldsymbol{M})^{\top})), \\
    \boldsymbol{G}_{\text{neg}} & = \operatorname{SoftMax}(\operatorname{ReLU}(\boldsymbol{E}_{\text{neg}}\boldsymbol{M}(\boldsymbol{E}_{\text{pos}}\boldsymbol{M})^{\top})),
\end{aligned}
\label{eq:interf}
\end{equation}
where $\boldsymbol{E}_{\text{pos}}\boldsymbol{M}(\boldsymbol{E}_{\text{neg}}\boldsymbol{M})^{\top}$ captures how the positive preference of each node is suppressed by the negative preference of other nodes. $\operatorname{ReLU}$ retains only effective (non-negative) interactions, and row-wise $\operatorname{SoftMax}$ helps to highlight the most influential ones via probability normalization.
By aggregating the interference over the interfering nodes with $\boldsymbol{u}_{\text{N}} = [\frac{1}{N}, \dots, \frac{1}{N}]^{\top}$ and $L_1$-normalizing the result ($\operatorname{Norm}(\boldsymbol{x}) = \boldsymbol{x}/\|\boldsymbol{x}\|_1$), we obtain a slow preference prior over nodes:
\begin{equation}
    \boldsymbol{p}_{\text{slow}} = \operatorname{Norm}\!\left((\boldsymbol{G}_{\text{pos}}^{\top}+\boldsymbol{G}_{\text{neg}}^{\top})\boldsymbol{u}_{\text{N}}\right),
\label{eq:slowprior}
\end{equation}
which summarizes the overall effect of the network on the preferences of each node.

In addition to the slow prior shared across samples, we account for the fast and sample-specific salience of each node. Let $\bar{\boldsymbol{H}}_{\text{node}}$ denote the global average of $\boldsymbol{H}_{\text{node}}$ across the node dimension. We define the node distinctiveness $\gamma_{\text{dist},i} = \|\boldsymbol{H}_{\text{node},i}-\bar{\boldsymbol{H}}_{\text{node}}\|_2$ and the memory confidence $\gamma_{\text{conf},i} = \max_k \boldsymbol{a}_{ik}$, where $\operatorname{z}(\cdot)$ standardizes a score across nodes (per sample) by removing its mean and dividing by its standard deviation. The two scores are standardized and aggregated into a contextual prior:
\begin{equation}
    \boldsymbol{p}_{\text{ctx}} = \operatorname{SoftMax}\!\left(\frac{\operatorname{z}(\boldsymbol{\gamma}_{\text{dist}}) + \tfrac{1}{2}\operatorname{z}(\boldsymbol{\gamma}_{\text{conf}})}{T_{\text{ctx}}}\right),
\label{eq:ctxprior}
\end{equation}
where $T_{\text{ctx}}$ is a smoothing temperature.
Finally, we obtain the node-prototype preference bias through a log-linear fusion of the slow and contextual priors, followed by a logarithmic compression:
\begin{equation}
    \boldsymbol{B} = \alpha \log\!\left(\operatorname{SoftMax}\!\left(\log(\boldsymbol{p}_{\text{slow}}) + \beta\log(\boldsymbol{p}_{\text{ctx}})\right)\right),
\label{eq:bias}
\end{equation}
where $\boldsymbol{B} \in \mathbb{R}^{N \times 1}$ serves as the resulting bias. $\alpha$ and $\beta$ are scaling coefficients.

\paragraph{Summary Tokens}
Given the node tokens $\boldsymbol{H}_{\text{node}}$ and the preference bias $\boldsymbol{B}$, we summarize node tokens into a compact set of tokens using learnable queries, following the Q-former paradigm \cite{li2022blip}. The tokenizer applies multi-head attention with $n_{\text{h}}$ heads in an internal dimension $d_{\text{t}}$ ($d_{\text{h}} = d_{\text{t}}/n_{\text{h}}$ per head); the standard head-wise and output projections inside multi-head attention are absorbed into the notations below for clarity. Specifically, we compute an assignment matrix that captures the soft association between summary tokens and node tokens, and aggregate node tokens based on the assignment matrix:
\begin{equation}
\begin{aligned}
    \boldsymbol{S}_{\text{sum}} & = \operatorname{SoftMax}\!\left(\frac{\boldsymbol{Q}_{\text{sum}}\boldsymbol{W}_{\text{q}}\left(\boldsymbol{H}_{\text{node}}\boldsymbol{W}_{\text{k}}\right)^{\top}
    }{\sqrt{d_{\text{h}}}}
    +\boldsymbol{B}^{\top}\right), \\
    \boldsymbol{T}_{\text{sum}} & = \operatorname{LayerNorm}\!\left(\boldsymbol{W}_{\text{o}}\boldsymbol{S}_{\text{sum}}(\boldsymbol{H}_{\text{node}}\boldsymbol{W}_{\text{v}})\right),
\end{aligned}
\label{eq:sumtok}
\end{equation}
where $\boldsymbol{Q}_{\text{sum}} \in \mathbb{R}^{N_{\text{sum}} \times d_{\text{t}}}$ represents the learnable queries, $\boldsymbol{W}_{\text{q}}, \boldsymbol{W}_{\text{k}}, \boldsymbol{W}_{\text{v}}$ map the inputs into the token space, and $\boldsymbol{W}_{\text{o}}$ maps the aggregated features back to the model dimension. $\boldsymbol{B}^{\top}$ is added, broadcasting along the summary token dimension. $\boldsymbol{S}_{\text{sum}} \in \mathbb{R}^{N_{\text{sum}} \times N}$ denotes the assignment weights. $\boldsymbol{T}_{\text{sum}} \in \mathbb{R}^{N_{\text{sum}} \times d}$ represents the resulting summary tokens.

\subsection{Memory-Calibrated Router}

\paragraph{Filtering Criteria}
Since summary tokens may omit informative details, we introduce a router to filter details with three criteria.

\textbf{Compression deficiency}. This metric quantifies the extent to which a node remains underrepresented by the summary tokens, and is measured by the summary coverage. The coverage is formulated as:
\begin{equation}
   \boldsymbol{\gamma}_{\text{C}} =\boldsymbol{S}_{\text{sum}}^{\top} \boldsymbol{u}_{\text{sum}},
\label{eq:cover}
\end{equation}
where $\boldsymbol{S}_{\text{sum}}$ (Eq. \eqref{eq:sumtok}) represents the assignment matrix. $\boldsymbol{u}_{\text{sum}} = [\frac{1}{N_{\text{sum}}}, \cdots, \frac{1}{N_{\text{sum}}}]^{\top}$ denotes a uniform distribution used to aggregate across summary tokens. $\boldsymbol{\gamma}_{\text{C}} \in \mathbb{R}^{N \times 1}$ represents the coverage scores. A higher value in $\boldsymbol{\gamma}_{\text{C}}$ signifies that a node's features are effectively compressed and, consequently, more reconstructible via the summary tokens.

\textbf{Contextual deviation}. This metric measures the extent to which each node's compression residual deviates from its neighborhood. We first reconstruct each node from the summary assignment and obtain the compression residual
\begin{equation}
    \boldsymbol{R} = \boldsymbol{H}_{\text{node}} - \operatorname{diag}(\boldsymbol{S}_{\text{sum}}^{\top}\boldsymbol{1})^{-1}\boldsymbol{S}_{\text{sum}}^{\top}\boldsymbol{S}_{\text{sum}}\boldsymbol{H}_{\text{node}},
\label{eq:residual}
\end{equation}
where $\boldsymbol{1} \in \mathbb{R}^{N_{\text{sum}}}$ is an all-ones vector. The deviation is formulated as:
\begin{equation}
    \boldsymbol{\gamma}_{\text{D}} = \| \boldsymbol{R} -\bar{\boldsymbol{A}}\boldsymbol{R}\|_2,
\label{eq:dev}
\end{equation}
where $\boldsymbol{\gamma}_{\text{D}} \in \mathbb{R}^{N \times 1}$ represents the deviation scores, the norm is computed row-wise over the node dimension, and $\bar{\boldsymbol{A}}$ is the symmetrized and row-normalized adjacency matrix. A higher value in $\boldsymbol{\gamma}_{\text{D}}$ indicates more distinctive node-level signals that stand out from the collective neighborhood dynamics and cannot be recovered from either the summaries or the neighbors.

\textbf{Prototype alignment}. This metric quantifies the consistency of current inputs with memory prototypes, helping to mitigate the impact of stochastic fluctuations. The alignment is formulated as:
\begin{equation}
\boldsymbol{\gamma}_{\text{A},i} = \boldsymbol{s}_{i}^{(1)} - \boldsymbol{s}_{i}^{(2)},
\label{eq:align}
\end{equation}
where $\boldsymbol{s}_{i}^{(1)}$ and $\boldsymbol{s}_{i}^{(2)}$ denote the largest and second-largest cosine similarities between the memory query $\boldsymbol{q}_i = \boldsymbol{H}_{\text{in},i}\boldsymbol{W}_{\text{a}}$ (Eq. \eqref{eq:memattn}) and the $K$ prototypes. The alignment scores $\boldsymbol{\gamma}_{\text{A}} \in \mathbb{R}^{N \times 1}$ are derived row-wise from the similarity matrix. A larger margin signifies strong and unambiguous adherence to a specific memory prototype, whereas a smaller margin suggests ambiguous matching, stochastic noise, or distribution shifts.

Based on the three complementary metrics, we evaluate the significance of each node's dynamics. The metrics are standardized across nodes and combined as:
\begin{equation}
    \boldsymbol{\gamma} = \operatorname{z}(\boldsymbol{\gamma}_{\text{D}}) + \operatorname{z}(\boldsymbol{\gamma}_{\text{A}}) - \operatorname{z}(\boldsymbol{\gamma}_{\text{C}}),
\label{eq:fusion}
\end{equation}
\begin{equation}
    \boldsymbol{\gamma}_{\text{F}}, \boldsymbol{I}_{\text{F}} = \operatorname{TopK} (\boldsymbol{\gamma}),
\label{eq:topk}
\end{equation}
where $\operatorname{z}(\cdot)$ standardizes each criterion across nodes within a sample. $\boldsymbol{I}_{\text{F}} \in \mathbb{R}^{K_{\text{F}} \times 1}$ denotes the indices of the selected nodes. $\boldsymbol{\gamma}_{\text{F}} \in \mathbb{R}^{K_{\text{F}} \times 1}$ represents their corresponding significance scores. The selection scores are detached from the gradients before top-$k$ selection to stabilize routing. This multi-criteria approach allows the model to prioritize the dynamics of reliable nodes, specifically those that are (1) locally informative, (2) strongly supported by memory prototypes, yet (3) insufficiently represented by summary tokens.

\paragraph{Detail Tokens}
Given the filtering indicator $\boldsymbol{I}_{\text{F}}$, we extract the corresponding node tokens $ \boldsymbol{H}_{\text{node}} ^{\star} \in \mathbb{R}^{K_{\text{F}} \times d}$ and significance scores $\boldsymbol{\gamma}_{\text{F}}$. To emphasize the dynamics of these reliable nodes, the standardized significance scores are injected as an attention bias:
\begin{equation}
    \boldsymbol{B}^{\star} = \alpha_{\text{r}}\operatorname{z}(\boldsymbol{\gamma}_{\text{F}}) \in \mathbb{R}^{K_{\text{F}} \times 1},
\label{eq:detailbias}
\end{equation}
where $\alpha_{\text{r}}$ is a scaling coefficient. The detail tokens are generated via attention, analogous to that of the summary tokens (Eq. \eqref{eq:sumtok}):
\begin{equation}
\begin{aligned}
    \boldsymbol{S}_{\text{detail}} & = \operatorname{SoftMax}\!\left(\frac{\boldsymbol{Q}_{\text{detail}}\boldsymbol{W}_{\text{q}}'\left(\boldsymbol{H}_{\text{node}}^{\star}\boldsymbol{W}_{\text{k}}'\right)^{\top}
    }{\sqrt{d_{\text{h}}}} +\boldsymbol{B}^{\star\top}\right), \\
    \boldsymbol{T}_{\text{detail}} & = \operatorname{LayerNorm}\!\left(\boldsymbol{W}_{\text{o}}'\boldsymbol{S}_{\text{detail}}(\boldsymbol{H}_{\text{node}}^{\star}\boldsymbol{W}_{\text{v}}')\right),
\end{aligned}
\label{eq:detailtok}
\end{equation}
where $\boldsymbol{T}_{\text{detail}} \in \mathbb{R}^{N_{\text{detail}} \times d}$ is the detail tokens. $\boldsymbol{S}_{\text{detail}} \in \mathbb{R}^{N_{\text{detail}} \times K_{\text{F}}}$ is the assignment matrix.

\subsection{LLM Fine-Tuning and Output Projection}
We follow the backbone selection and fine-tuning strategies of prior works \cite{stllm,Std-plm,onefitsall}. Specifically, we fully update the small modules (i.e., layer normalization and positional embeddings) to align the LLM with input statistics, and apply LoRA \cite{lora} to the attention layers for adaptation while keeping the feed-forward networks frozen.

To provide the LLM with a compact view of the global temporal evolution, we construct two time tokens from the network-averaged observations before the summary and detail tokens. Let $\bar{\boldsymbol{x}}^{(t)} \in \mathbb{R}^{F}$ denote the network-averaged observations at input step $t$, and $\Delta\bar{\boldsymbol{x}}^{(t)} = \bar{\boldsymbol{x}}^{(t+1)}-\bar{\boldsymbol{x}}^{(t)}$ its temporal difference. The time tokens are computed as:
\begin{equation}
\begin{aligned}
    \boldsymbol{t}_{\text{state}} & = \operatorname{MLP}_{\text{s}}(\bar{\boldsymbol{x}}^{(1)} \| \cdots \| \bar{\boldsymbol{x}}^{(T)} \| \boldsymbol{e}_{\text{t}}^{(T)}), \\
    \boldsymbol{t}_{\text{trend}} & = \operatorname{MLP}_{\text{g}}(\Delta\bar{\boldsymbol{x}}^{(1)} \| \cdots \| \Delta\bar{\boldsymbol{x}}^{(T-1)} \| \boldsymbol{e}_{\text{t}}^{(T)}), \\
    \boldsymbol{T}_{\text{time}} & = \operatorname{LayerNorm}([\boldsymbol{t}_{\text{state}};\, \boldsymbol{t}_{\text{trend}}]),
\end{aligned}
\label{eq:timetok}
\end{equation}
where $\boldsymbol{T}_{\text{time}} \in \mathbb{R}^{2 \times d}$ consists of a state token and a trend token, and $\boldsymbol{e}_{\text{t}}^{(T)}$ is the temporal embedding of the last input step.

We concatenate the time, summary, and detail tokens, and feed them into the LLM:
\begin{equation}
    \boldsymbol{O}_{\text{LLM}} = \operatorname{LLM}(\boldsymbol{T}_{\text{time}} \| \boldsymbol{T}_{\text{sum}} \| \boldsymbol{T}_{\text{detail}}),
\label{eq:llm}
\end{equation}
where $\boldsymbol{O}_{\text{LLM}} \in \mathbb{R}^{(N_{\text{sum}}+N_{\text{detail}}+2) \times d_{\text{LLM}}}$ denotes the hidden states from the LLM's last layer, serving as the final encoded representations of the inputs. Consequently, these representations are split into the time, summary, and detail parts, i.e., $\boldsymbol{O}_{\text{sum}} \in \mathbb{R}^{N_{\text{sum}} \times d_{\text{LLM}}}$ and $\boldsymbol{O}_{\text{detail}} \in \mathbb{R}^{N_{\text{detail}} \times d_{\text{LLM}}}$.

These representations are distributed back to each node via multi-head cross-attention decoding, where the node tokens serve as queries, the corresponding learnable queries serve as keys, and the hidden states serve as values:
\begin{equation}
\begin{aligned}
    \boldsymbol{D}_{\text{sum}} & = \operatorname{Dec}(\boldsymbol{H}_{\text{node}},\, \boldsymbol{Q}_{\text{sum}},\, \boldsymbol{O}_{\text{sum}}), \\
    \boldsymbol{D}_{\text{detail}}^{\star} & = \operatorname{Dec}(\boldsymbol{H}_{\text{node}}^{\star},\, \boldsymbol{Q}_{\text{detail}},\, \boldsymbol{O}_{\text{detail}}),
\end{aligned}
\label{eq:dec}
\end{equation}
where $\operatorname{Dec}(\boldsymbol{Q},\boldsymbol{K},\boldsymbol{V})$ denotes the same multi-head cross-attention decoding scheme as the tokenizer, i.e., sinusoidal positional encodings on queries and keys with a final projection and layer normalization.

The decoded representations are fused into node-wise features:
\begin{equation}
    \boldsymbol{Z} = \boldsymbol{D}_{\text{sum}} + \operatorname{Scatter}(\eta\, \boldsymbol{D}_{\text{detail}}^{\star}) + \boldsymbol{c}_{\text{trend}},
\label{eq:nodefeat}
\end{equation}
where $\operatorname{Scatter}(\cdot)$ scatters the detail features back to the nodes indexed by $\boldsymbol{I}_{\text{F}}$ and pads the unmapped nodes with zeros, $\eta = \delta \sum_{i \in \boldsymbol{I}_{\text{F}}} \operatorname{SoftMax}(\boldsymbol{\gamma})_i$ is a routed-confidence factor scaling the detail correction ($\delta$ being a residual scale and $\boldsymbol{\gamma}$ the node significance in \eqref{eq:fusion}), and $\boldsymbol{c}_{\text{trend}} = \boldsymbol{O}_{\text{trend}} + \boldsymbol{t}_{\text{trend}}$ is the temporal context of the trend time token, broadcast to all nodes to inject the global temporal evolution. The fused features then refine the node tokens in a residual manner and are projected to the target horizon:
\begin{equation}
    \hat{\boldsymbol{X}}^{T+1:T+P} = \operatorname{MLP}\!\left(\operatorname{LayerNorm}\big(\boldsymbol{Z} + \boldsymbol{H}_{\text{node}}\big)\right),
\label{eq:pred}
\end{equation}
where $\hat{\boldsymbol{X}}^{T+1:T+P} \in \mathbb{R}^{N \times P}$ denotes the final predictions.

\subsection{Training Loss}
The training objective integrates a forecasting loss with regularization terms for the memory prototypes and the summary assignments.
Following prior works \cite{timellm,fstllm}, we adopt MAE over the prediction horizon as the forecasting loss.
To regularize the memory, we introduce an InfoNCE objective to contrast the most relevant prototype for each node against all other entries in the memory bank. Specifically, let $\boldsymbol{q}_i = \boldsymbol{H}_{\text{in},i} \boldsymbol{W}_{\text{a}}$ denote the memory query derived from the input. Based on the attention weights $\boldsymbol{a}$ (Eq. \eqref{eq:memattn}), we identify the most relevant prototype as $\boldsymbol{m}_i^+$. This regularization term enhances retrieval discriminability by pulling the query closer to its corresponding prototype while pushing it away from irrelevant entries. The losses are defined as:
\begin{equation}
\begin{aligned}
\mathcal{L}_{\text{pred}}
& =
\frac{1}{N P}
\sum_{i=1}^{N}
\sum_{t=1}^{P}
\left\|
\hat{\boldsymbol{X}}_{i,t} - \boldsymbol{X}_{i,t}
\right\|_{1} ,\\
\mathcal{L}_{\text{nce}}
&  =
-\frac{1}{N}\sum_{i=1}^{N}
\log
\frac{
\exp(\cos(\boldsymbol{q}_i,\boldsymbol{m}_i^+)/\tau)
}{
\sum_{k=1}^{K}
\exp(\cos(\boldsymbol{q}_i,\boldsymbol{m}_k)/\tau)
},
\end{aligned}
\label{eq:lpredlnce}
\end{equation}
where $\cos(\cdot,\cdot)$ denotes cosine similarity, consistent with the memory attention (Eq. \eqref{eq:memattn}). $N$ denotes the number of nodes and $P$ is the prediction horizon. $\tau$ is the shared temperature parameter and is set to 0.2 in all experiments.

Beyond the memory regularization, we shape the summary coverage distribution toward a degree-aware prior via a Dirichlet likelihood. Since $\boldsymbol{S}_{\text{sum}}^{\top}\boldsymbol{u}_{\text{sum}}$ measures the total attention each node receives from the summary tokens, we define:
\begin{equation}
\mathcal{L}_{\text{dir}}
=
-\sum_{b=1}^{B}
\log p_{\text{Dir}}
\left(
\operatorname{SoftMax}(\boldsymbol{S}_{\text{sum}}^{(b)\top}\boldsymbol{u}_{\text{sum}})
\,\middle|\,
\boldsymbol{\alpha}
\right),
\label{eq:ldir}
\end{equation}
where $p_{\text{Dir}}(\cdot|\boldsymbol{\alpha})$ denotes the Dirichlet density with concentration parameters $\boldsymbol{\alpha} = 1.05 + 5\,\operatorname{SoftMax}(\boldsymbol{d})$ conditioned on the node degrees $\boldsymbol{d}$, encouraging well-connected nodes to attract more summary attention.

The total loss is defined as:
\begin{equation}
\mathcal{L}
=
\mathcal{L}_{\text{pred}}
+
\lambda_{\text{nce}}\mathcal{L}_{\text{nce}}
+
\mathcal{L}_{\text{dir}},
\label{eq:total}
\end{equation}
where $\lambda_{\text{nce}}$ controls the trade-off between forecasting and memory regularization, and its weight is linearly warmed up over the first $E_{\text{w}}$ epochs.

\section{Experiments}
\subsection{Experimental Protocol}
\textbf{Datasets.}
We conduct comprehensive evaluations of STMemoLLM on eight real-world benchmarks spanning four traffic forecasting tasks: (1) traffic flow forecasting on PEMS03, PEMS04, PEMS07, and PEMS08, (2) traffic speed forecasting on METR-LA and PEMS-BAY, (3) taxi demand forecasting on NYCTaxi, and (4) bike-sharing demand forecasting on CHBike. The dataset statistics are summarized in Table~\ref{tab:dataset_statistics}.
 
\begin{table}[]
    \centering
    \caption{Dataset statistics.}
    \label{tab:dataset_statistics}
    \resizebox{0.5\textwidth}{!}{
    \begin{tabular}{lcccc}
        \toprule
        Dataset & \#Nodes & Time Range & Granularity & Type \\
        \midrule
        PEMS03 & 358 & 9/1/2018--11/30/2018 & 5 min & Traffic Flow \\
        PEMS04 & 307 & 1/1/2018--2/28/2018 & 5 min & Traffic Flow \\
        PEMS07 & 883 & 5/1/2017--8/31/2017 & 5 min & Traffic Flow \\
        PEMS08 & 170 & 7/1/2016--8/31/2016 & 5 min & Traffic Flow \\
        METR-LA & 207 & 3/1/2012--6/30/2012 & 5 min & Traffic Speed \\
        PEMS-BAY & 325 & 1/1/2017--5/31/2017 & 5 min & Traffic Speed \\
        NYCTaxi & 266 & 4/1/2016--6/30/2016 & 30 min & Taxi Demand \\
        CHBike & 250 & 4/1/2016--6/30/2016 & 30 min & Bike Demand \\
        \bottomrule
    \end{tabular}}
\end{table}

Specifically, PEMS03, PEMS04, PEMS07, and PEMS08 are obtained from Song et al.~\cite{stsgcn}. These datasets are derived from the California Transportation Agencies (CalTrans) Performance Measurement System (PeMS)~\footnote{http://pems.dot.ca.gov/}, collecting traffic flow from sensors deployed across major freeway networks in four metropolitan districts in California. METR-LA and PEMS-BAY are obtained from Li et al.~\cite{dcrnn}. METR-LA is derived from the Los Angeles County Metropolitan Transportation Authority~\footnote{https://www.metro.net/}, collecting traffic speed from loop detectors in the highways of Los Angeles County. PEMS-BAY is constructed from CalTrans PeMS, collecting traffic speed from sensors deployed across freeway networks in the Bay Area. NYCTaxi and CHBike are obtained from Liu et al.~\cite{liu2025st}. NYCTaxi is derived from the Taxi \& Limousine Commission of the City of New York~\footnote{https://www.nyc.gov/site/tlc/about/tlc-trip-record-data.page}, collecting taxicab trip records from TLC Trip Record Data. CHBike is derived from the bike-sharing system \footnote{https://citibikenyc.com/}, collecting Citi bike orders.

We follow the established preprocessing protocols from the original dataset studies. For the traffic flow and traffic speed datasets, the data are aggregated to 5-minute intervals. For the taxi and bike demand datasets, the orders are clustered into stations, with each station characterized by two features: pick-up demand and drop-off demand. All datasets are chronologically split into training, validation, and test sets following the original studies. For all datasets, we fill the missing data by linear interpolation and normalize the datasets using Z-Score normalization.

\textbf{Baselines.}
\begin{table*}
    \centering
    \caption{Results on PEMS03/04/07/08.}
    \label{tab:pems-results}
    \resizebox{0.8 \textwidth}{!}{
    \begin{tabular}{c|ccc|ccc|ccc|ccc}
        \toprule
        \multirow{2}{*}{Method} & \multicolumn{3}{c|}{PEMS03} & \multicolumn{3}{c|}{PEMS04} & \multicolumn{3}{c|}{PEMS07} & \multicolumn{3}{c}{PEMS08} \\
        \cmidrule(lr){2-4} \cmidrule(lr){5-7} \cmidrule(lr){8-10} \cmidrule(lr){11-13}
        & MAE & RMSE & MAPE & MAE & RMSE & MAPE & MAE & RMSE & MAPE & MAE & RMSE & MAPE \\
        \midrule
        STGCN (2018) & 17.55 & 30.42 & 17.34 & 21.16 & 34.89 & 13.83 & 25.33 & 39.34 & 11.21 & 17.50 & 27.09 & 11.29 \\
        ASTGCN (2019) & 17.34 & 29.56 & 17.21 & 22.93 & 35.22 & 16.56 & 24.01 & 37.87 & 10.73 & 18.25 & 28.06 & 11.64 \\
        GWaveNet (2019) & 19.12 & 32.77 & 18.89 & 24.89 & 39.66 & 17.29 & 26.39 & 41.50 & 11.97 & 18.28 & 30.05 & 12.15 \\
        AGCRN (2020) & 15.98 & 28.25 & 15.23 & 19.83 & 32.26 & 12.97 & 22.37 & 36.55 & 9.12 & 15.95 & 25.22 & 10.09 \\
        MTGNN (2020) & 15.10 & 25.93 & 15.67 & 19.32 & 31.57 & 13.52 & 22.07 & 35.80 & 9.21 & 15.71 & 24.62 & 10.03 \\
        STSGCN (2020) & 17.48 & 29.21 & 16.78 & 21.19 & 33.65 & 13.90 & 24.26 & 39.03 & 10.21 & 17.13 & 26.80 & 10.96 \\
        STFGNN (2021) & 16.77 & 28.34 & 16.30 & 20.48 & 32.51 & 16.77 & 23.46 & 36.60 & 9.21 & 16.94 & 26.25 & 10.60 \\
        DGCRN (2023) & 15.21$^*$ & 26.86$^*$ & 15.88$^*$ & 19.25$^*$ & 31.17$^*$ & 12.66$^*$ & 22.21$^*$ & 35.85$^*$ & 9.15$^*$ & 15.98$^*$ & 25.67$^*$ & 9.97$^*$ \\
        PDFormer (2023) & 14.74 & 25.59 & 15.35 & 18.31 & 29.97 & 12.10 & 19.83 & 32.87 & 8.53 & 13.58 & 23.51 & 9.05 \\
        MegaCRN (2023) & 14.91 & 25.96 & 15.50 & 18.78 & 30.67 & 13.11 & 19.85 & 32.71 & 8.38 & 16.17 & 25.26 & 10.87 \\
        STDN (2025) & 15.65$^*$ & 27.72$^*$ & 16.64$^*$ & 18.40 & 30.41 & 12.21 & 20.08 & 33.73 & 9.29 & 14.71$^*$ & 24.44$^*$ & 10.50$^*$ \\
        ST-SSDL (2025) & 14.68 & 25.60 & 15.67 & \underline{18.08} & \textbf{29.64} & 12.36 & \underline{19.19} & \underline{32.51} & 8.11 & 13.86 & \underline{23.10} & 9.17 \\
        HyperD (2026) & 14.69$^*$ & \textbf{23.66}$^*$ & 15.76$^*$ & 18.20$^*$ & 29.94$^*$ & 12.44$^*$ & 19.37$^*$ & 33.22$^*$ & \underline{8.05}$^*$ & 13.59$^*$ & 23.24$^*$ & 8.91$^*$ \\
        M$^2$Net (2027) & 14.65$^*$ & 24.68$^*$ & 15.17$^*$ & 18.20$^*$ & \underline{29.89}$^*$ & 12.43$^*$ & 19.39$^*$ & 32.52$^*$ & 8.29$^*$ & 14.02$^*$ & \underline{23.10}$^*$ & 9.15$^*$ \\
        \midrule
        LagLLaMA (2023) & 16.38$^*$ & 27.64$^*$ & 17.21$^*$ & 23.40$^*$ & 36.09$^*$ & 21.54$^*$ & 24.47$^*$ & 38.96$^*$ & 15.15$^*$ & 15.91$^*$ & 25.00$^*$ & 18.34$^*$ \\
        STGLLM (2024) & 15.26 & \underline{24.11} & 15.73 & 20.00 & 32.11 & 13.69 & 21.98 & 35.02 & 9.72 & 15.53 & 24.74 & 10.15 \\
        STLLM (2024) & 17.25 & 27.25 & 22.96 & 19.00 & 30.35 & 13.55 & 21.48 & 34.07 & 10.20 & 14.67 & 23.50 & 10.63 \\
        TimeCMA (2025) & 16.74$^*$ & 25.37$^*$ & 16.88$^*$ & 21.79$^*$ & 32.68$^*$ & 15.01$^*$ & 25.51$^*$ & 38.60$^*$ & 12.91$^*$ & 17.82$^*$ & 26.90$^*$ & 11.19$^*$ \\
        CrossST (2025) & 15.73$^*$ & 24.98$^*$ & 19.12$^*$ & 20.41$^*$ & 32.16$^*$ & 14.11$^*$ & 22.30$^*$ & 34.66$^*$ & 13.36$^*$ & 16.02$^*$ & 24.46$^*$ & 13.49$^*$ \\
        STLLM+ (2025) & 15.91$^*$ & 27.33$^*$ & 17.20$^*$ & 18.58$^*$ & 30.46$^*$ & 13.40$^*$ & 20.92$^*$ & 34.82$^*$ & 9.28$^*$ & 14.68$^*$ & 23.87$^*$ & 10.47$^*$ \\
        ST-LINK (2025) & 14.83 & 26.27 & \underline{14.90} & 18.68 & 30.29 & 12.83 & 20.13 & 33.12 & 8.53 & 14.87 & 23.93 & 9.76 \\
        STD-PLM (2025) & \underline{14.59} & 25.36 & 14.92 & 18.16 & 30.21 & \underline{11.89} & 19.25 & 32.84 & 8.06 & \underline{13.31} & 23.19 & \underline{8.84} \\
        \midrule
        STMemoLLM & \textbf{14.44} & 24.93 & \textbf{14.65} & \textbf{18.01} & 30.04 & \textbf{11.71} & \textbf{18.80} & \textbf{32.50} & \textbf{7.87} & \textbf{13.16} & \textbf{23.08} & \textbf{8.75} \\
        \bottomrule
    \end{tabular}}
\end{table*}
We compare STMemoLLM with 22 end-to-end and pretrained-based baselines.
(1)~\emph{End-to-end models}:
STGCN~\cite{stgcn} combines graph convolutions over the road network with gated temporal convolutions, jointly capturing spatial and temporal dependencies in a unified framework.
ASTGCN~\cite{astgcn} further equips this paradigm with spatial and temporal attention, thus dynamically emphasizing spatial correlations and multi-period temporal patterns.
GraphWaveNet~\cite{graph_wavenet} learns a self-adaptive adjacency matrix from trainable node embeddings and pairs it with dilated causal temporal convolutions, removing the reliance on a predefined graph.
AGCRN~\cite{agcrn} parameterizes the adaptive adjacency matrix and node-specific transformation weights with node embeddings, and integrates them into a GRU-style recurrent architecture to capture node-level spatial-temporal correlations.
MTGNN~\cite{mtgnn} infers inter-variable dependencies from data via a graph learning module and couples mix-hop graph convolutions with dilated inception temporal convolutions.
STSGCN~\cite{stsgcn} constructs localized spatial-temporal graphs that connect each node with its neighbors across adjacent time steps, and applies spatial-temporal synchronous graph convolutions to capture localized heterogeneous correlations.
STFGNN~\cite{stfgcn} merges the spatial graph with temporal-connectivity graphs into a spatial-temporal fusion graph, over which graph convolutions and gated temporal convolutions jointly model both types of dependencies.
DGCRN~\cite{dgcrn} dynamically generates node embeddings and the corresponding graph structure conditioned on the current input at each recurrent step, embedding them into a graph convolutional GRU to capture input-dependent spatial correlations.
PDFormer~\cite{pdformer} is a Transformer-based predictor that encodes graph structure into spatial attention via graph masking, and introduces time-delay-aware spatial attention to explicitly model long-range dependencies and propagation delays.
MegaCRN~\cite{megacrn} augments a graph convolutional recurrent encoder--decoder with a learnable pattern memory and a meta-graph learner that generates memory-conditioned node embeddings for adaptive graph construction.
STDN~\cite{STDN} learns spatio-temporal embeddings and a dynamic graph that captures high-order node interactions, and refines node representations with a spatiotemporal-aware trend-seasonality decomposition module.
ST-SSDL~\cite{gao2025how} anchors each input to its timestamp-aligned historical average and discretizes the latent space with learnable prototypes of typical spatio-temporal patterns; contrastive and deviation losses regularize the latent structure to quantify how far the current input deviates from history.
HyperD~\cite{shao2026hyperd} decouples traffic data into periodic and residual components, extracting daily and weekly patterns with learnable periodic embeddings and spatial-temporal attention.
M$^2$Net~\cite{zhou2027m2net} unifies scale-specific spatial structures with prototype decomposition, separating shared patterns from node-specific residuals for spatial-temporal forecasting.

\begin{table*}
    \centering
    \caption{Results on METR-LA, PEMS-BAY, NYCTaxi, and CHBike.}
    \label{tab:urban-results}
    \resizebox{\textwidth}{!}{
    \begin{tabular}{l|ccc|ccc|ccc|ccc|ccc|ccc}
        \toprule
        \multirow{2}{*}{Method} & \multicolumn{3}{c|}{\multirow{2}{*}{METR-LA}} & \multicolumn{3}{c|}{\multirow{2}{*}{PEMS-BAY}} & \multicolumn{6}{c|}{NYCTaxi} & \multicolumn{6}{c}{CHBike} \\
        \cmidrule(lr){8-13} \cmidrule(lr){14-19}
        & \multicolumn{3}{c|}{} & \multicolumn{3}{c|}{} & \multicolumn{3}{c|}{Pick-up} & \multicolumn{3}{c|}{Drop-off} & \multicolumn{3}{c|}{Pick-up} & \multicolumn{3}{c}{Drop-off} \\
        \midrule
        Metric & MAE & RMSE & MAPE & MAE & RMSE & MAPE & MAE & RMSE & MAPE & MAE & RMSE & MAPE & MAE & RMSE & MAPE & MAE & RMSE & MAPE \\
        \midrule
        STLLM (2024) & 3.13$^*$ & 6.37$^*$ & \underline{8.54}$^*$ & 1.60$^*$ & \textbf{3.58}$^*$ & 3.67$^*$ & 5.29 & 9.42 & 35.41 & 5.07 & 9.07 & \underline{34.59} & 2.07 & 3.23 & 58.93 & 1.91 & 2.83 & 51.71 \\
        CrossST (2025) & 3.17$^*$ & 6.42$^*$ & 8.75$^*$ & 1.75$^*$ & 4.25$^*$ & 4.10$^*$ & 5.58$^*$ & 9.78$^*$ & 38.14$^*$ & 5.29$^*$ & 9.60$^*$ & 37.97$^*$ & 2.23$^*$ & 3.46$^*$ & 60.13$^*$ & 2.08$^*$ & 3.11$^*$ & 55.82$^*$ \\
        STLLM+ (2025) & 3.17$^*$ & 6.48$^*$ & 8.66$^*$ & 1.68$^*$ & 3.71$^*$ & 3.77$^*$ & 5.18 & 8.98 & \underline{33.22} & 4.94 & 8.68 & 36.09 & 1.98 & 3.05 & 55.59 & 1.89 & 2.79 & 51.48 \\
        STD-PLM (2025) & \underline{3.11} & 6.34 & 8.59 & 1.65 & 3.64 & 4.59 & 5.07 & 8.86 & 35.76 & \underline{4.82} & \underline{8.46} & 35.63 & 1.66 & 2.79 & 58.61 & 1.57 & 2.53 & 57.00 \\
        ST-SSDL (2025) & 3.37 & 7.17 & 9.48 & 1.86 & 4.34 & 4.35 & 5.06 & \underline{8.75} & 33.90 & 4.88 & 8.59 & 34.87 & 2.24 & 3.50 & 65.33 & 1.87 & 2.78 & \textbf{50.44} \\
        ST-LINK (2025) & 3.13 & \underline{6.32} & 8.60 & 1.62 & \underline{3.60} & 3.64 & 5.10 & 8.86 & 33.29 & \underline{4.82} & 8.49 & 34.80 & 1.99 & 3.07 & \underline{51.78} & 1.87 & 2.85 & 52.33 \\
        HyperD (2026) & 3.40 & 6.69 & 10.38 & 1.72 & 4.01 & 4.17 & 5.09 & 8.86 & 36.06 & 4.91 & 8.64 & 37.22 & 1.79 & 3.23 & 55.64 & 1.70 & 2.91 & 53.79 \\
        M$^2$Net (2027) & 3.20 & 6.39 & 9.13 & \textbf{1.54} & \underline{3.60} & \underline{3.55} & \underline{5.01} & 8.88 & 35.54 & 4.84 & 8.75 & 35.18 & \underline{1.63} & \underline{2.77} & 58.50 & \underline{1.55} & \underline{2.52} & 56.44 \\
        \midrule
        STMemoLLM & \textbf{3.07} & \textbf{6.29} & \textbf{8.50} & \underline{1.56} & \textbf{3.58} & \textbf{3.47} & \textbf{4.99} & \textbf{8.73} & \textbf{33.10} & \textbf{4.76} & \textbf{8.38} & \textbf{34.13} & \textbf{1.62} & \textbf{2.72} & \textbf{50.11} & \textbf{1.54} & \textbf{2.47} & \underline{51.06} \\
        \bottomrule
    \end{tabular}}
\end{table*}
(2)~\emph{Pretrained-based models}:
LagLLaMA~\cite{lagllama} is a decoder-only time series foundation model pretrained on large and diverse corpora, enabling zero-shot probabilistic forecasting.
STGLLM~\cite{liu2024can} compresses graph-structured data into compact tokens via a spatial-temporal graph tokenizer and bridges them to a largely frozen LLM with a lightweight linear adapter.
STLLM~\cite{stllm} takes timesteps at each location as tokens, learns spatial-temporal embeddings that are fused into every token, and adapts the LLM through a partially frozen attention strategy.
TimeCMA~\cite{timecma} separately encodes numerical time series and textual prompts containing the same observations, and aligns the resulting representations through channel-wise similarity-based retrieval to support forecasting.
CrossST~\cite{crossst} pre-trains on large-scale cross-district datasets and stores universal spatio-temporal patterns in a pattern bank, combining frequency- and time-domain analysis for temporal dependencies with graph attention for dynamic spatial propagation.
STLLM+~\cite{liu2025st} extends STLLM by injecting a proximity-based adjacency matrix of the traffic network into the calibrated LLM via a partially frozen graph attention module, and adopts LoRA-augmented training to reduce computational overhead.
ST-LINK~\cite{jeon2025st} is an LLM-based spatio-temporal forecaster that combines Spatially-Enhanced Attention, which injects spatial relations into rotary position embeddings, with a Memory Retrieval Feed-Forward Network that retrieves historical patterns for temporal modeling.
STD-PLM~\cite{Std-plm} feeds explicitly designed spatial and temporal tokenizers together with topology-aware node embeddings into a PLM, and captures non-pairwise correlations with a sandglass attention module.

In particular, PDFormer~\cite{pdformer}, MegaCRN~\cite{megacrn}, and ST-SSDL~\cite{gao2025how} are memory-based baselines; and STGLLM~\cite{liu2024can}, STLLM~\cite{stllm}, STLLM+~\cite{liu2025st}, STD-PLM~\cite{Std-plm}, and ST-LINK~\cite{jeon2025st} are LLM-based baselines for spatial-temporal forecasting.

\textbf{Settings.}
All experiments are conducted on a single NVIDIA RTX 5090 GPU with 32GB of memory. We employ the Adam optimizer with an initial learning rate of $10^{-3}$. The batch size is set to 64, and the model is trained for up to 300 epochs with an early stopping patience of 10.
We split the four PeMS flow datasets chronologically into training, validation, and test segments with ratios 0.6/0.2/0.2. For METR-LA and PEMS-BAY, the released scripts use ratios 0.7/0.1/0.2. The taxi and bike experiments use the supplied train/validation/test NPZ files without re-splitting their samples. Input and prediction lengths are both 12 steps for every task. These correspond to one hour each for 5-minute data and six hours each for 30-minute data. 
We evaluate model performance using mean absolute error (MAE), root mean squared error (RMSE), and mean absolute percentage error (MAPE). Hyperparameters are tuned on the validation set with an AutoML toolkit NNI (Neural Network Intelligence) \cite{nni} to reduce computational costs. The joint validation search considers $K \in \{32,64,96,128,256\}$, $K_{\text{F}} \in \{12,24,32,48,96\}$, and the complete token configurations $(T,N_{\text{sum}},N_{\text{detail}}) \in \{(128,96,32),(128,112,16),(96,64,32),(96,72,24)\}$. Thus, the evaluated summary/detail ratios are $\rho\in\{2{:}1,3{:}1,7{:}1\}$.
For the full model, the memory contrastive loss uses weight $\lambda_{\text{nce}}=0.001$ and temperature $\tau=0.2$, with the contrastive weight linearly warmed up over the first 10 epochs.

\begin{table}
    \centering
    \caption{Five-seed comparison between STD-PLM and STMemoLLM.}
    \label{tab:multiseed}
    \resizebox{0.8 \linewidth}{!}{
    \begin{tabular}{c|c|cc}
        \toprule
        Dataset & Metric & STD-PLM & STMemoLLM \\
        \midrule
        PEMS03 & MAE & $14.576\pm0.009$ & $\mathbf{14.440\pm0.016}^{\dagger}$ \\
        & RMSE & $25.466\pm0.043$ & $\mathbf{24.951\pm0.057}^{\dagger}$ \\
        & MAPE (\%) & $14.938\pm0.011$ & $\mathbf{14.666\pm0.023}^{\dagger}$ \\
        PEMS04 & MAE & $18.160\pm0.012$ & $\mathbf{18.008\pm0.013}^{\dagger}$ \\
        & RMSE & $30.316\pm0.036$ & $\mathbf{30.122\pm0.037}^{\dagger}$ \\
        & MAPE (\%) & $11.928\pm0.033$ & $\mathbf{11.700\pm0.024}^{\dagger}$ \\
        PEMS07 & MAE & $19.300\pm0.036$ & $\mathbf{18.850\pm0.052}^{\dagger}$ \\
        & RMSE & $32.796\pm0.044$ & $\mathbf{32.544\pm0.052}^{\dagger}$ \\
        & MAPE (\%) & $8.102\pm0.023$ & $\mathbf{7.880\pm0.020}^{\dagger}$ \\
        PEMS08 & MAE & $13.318\pm0.022$ & $\mathbf{13.174\pm0.011}^{\dagger}$ \\
        & RMSE & $23.242\pm0.041$ & $\mathbf{23.098\pm0.022}^{\dagger}$ \\
        & MAPE (\%) & $8.866\pm0.011$ & $\mathbf{8.780\pm0.017}^{\dagger}$ \\
        \bottomrule
    \end{tabular}}
\end{table}

\subsection{Overall Performance}
Table \ref{tab:pems-results} and Table \ref{tab:urban-results} summarize the forecasting results of STMemoLLM and baselines. 
We highlight the best and second-best results in \textbf{bold} and \underline{underline}, respectively. Entries marked with $*$ denote results reproduced using official implementations under our settings, while the remaining results are taken from the original publications or related work \cite{Std-plm}. Several observations can be drawn: 
(1) STMemoLLM achieves the best or tied-best performance in 26 out of 30 metric comparisons, with an average relative MAE reduction of 1.92\% over STD-PLM across ten forecasting tasks on eight datasets. To assess variability across random seeds, we additionally repeat training for STMemoLLM and STD-PLM with five matched random seeds on PEMS03, PEMS04, PEMS07, and PEMS08. Table~\ref{tab:multiseed} reports the mean and sample standard deviation over the five runs. A superscript $\dagger$ indicates a significant improvement under a two-sided paired $t$-test with Holm correction across the 12 dataset--metric comparisons (Holm-adjusted $p<0.05$). Compared with STD-PLM, STMemoLLM reduces the mean MAE by 0.136, 0.152, 0.450, and 0.144 on PEMS03, PEMS04, PEMS07, and PEMS08, respectively, with the largest reduction among these four datasets observed on PEMS07.
(2) STMemoLLM achieves the best MAPE across all PEMS datasets, showing strong performance in low-magnitude regimes. Unlike standard LLMs that struggle with zero-inflated data, the memory-enhanced tokenizer provides stable prototypical anchors that reduce relative errors and prevent drift in sparse intervals. However, the relatively low MAE and higher RMSE suggest that it may be occasionally sensitive to extreme cases (e.g., sudden traffic spikes and rare congestion events). (3) STMemoLLM achieves the best results on all six metrics for NYCTaxi and on five of the six metrics for CHBike. For CHBike drop-off MAPE, it achieves the second-best value of 51.06\%, compared with the best value of 50.44\% from ST-SSDL. These results demonstrate competitive forecasting performance for both pick-up and drop-off demand. These may be characterized by discrete and high-variance demand patterns. The memory-calibrated router filters noise by matching residuals to historical prototypes, preserving consistent demand signals while suppressing non-prototypical fluctuations.

\subsection{Robustness Evaluation}
We evaluate the robustness of STMemoLLM under three settings: inference-time point corruption, structured sensor missingness, and limited supervision for few-shot settings. All models use the same data split, preprocessing procedure, and prediction horizon.

\textbf{Robustness to corruption.}
We evaluate robustness on PEMS04 under two types of inference-time
corruption: point masking and additive Gaussian noise. For point masking,
we randomly replace $r\in\{5\%,10\%,20\%\}$ of the node--time
observations in each input window with zero, corresponding to the
training mean under the normalization used. For Gaussian noise, we add
independent noise $\epsilon\sim\mathcal{N}(0,\sigma^2)$ to each normalized
input value, with $\sigma\in\{0.05,0.10,0.20\}$. All models receive the
same corrupted inputs and are evaluated without retraining or test-time
adaptation.

Table~\ref{tab:robustness-baseline} reports MAE and its relative increase
over each model's clean-input result. STMemoLLM achieves the lowest MAE
across all six corruption settings. Under point masking, it also exhibits
smaller relative MAE increases than STD-PLM at all three masking ratios.
Its absolute MAE advantage over STD-PLM grows from $0.26$ at 5\% masking
to $0.80$ at 20\% masking, indicating that the advantage persists and
becomes larger as more observations are masked.

\begin{table}[t]
    \centering
    \caption{MAE results under controlled corruption.}
    \label{tab:robustness-baseline}
    \resizebox{\linewidth}{!}{
    \begin{tabular}{l|ccc}
        \toprule
        Input condition & STMemoLLM & STD-PLM & MegaCRN \\
        \midrule
        Clean & 18.01 & 18.15 & 18.52 \\
        Point mask 5\% & 20.15 (+11.9\%) & 20.41 (+12.5\%) & 20.61 (+11.3\%) \\
        Point mask 10\% & 22.49 (+24.9\%) & 22.89 (+26.1\%) & 23.48 (+26.8\%) \\
        Point mask 20\% & 27.70 (+53.9\%) & 28.50 (+57.0\%) & 33.04 (+78.4\%) \\
        Gaussian noise $\sigma=0.05$ & 18.24 (+1.3\%) & 18.42 (+1.5\%) & 18.78 (+1.4\%) \\
        Gaussian noise $\sigma=0.10$ & 18.81 (+4.5\%) & 18.98 (+4.6\%) & 19.47 (+5.1\%) \\
        Gaussian noise $\sigma=0.20$ & 20.53 (+14.0\%) & 20.77 (+14.4\%) & 22.06 (+19.1\%) \\
        \bottomrule
    \end{tabular}}
\end{table}

\textbf{Robustness to missingness.}
We evaluate robustness on PEMS04 under two missingness patterns:
sensor-level and neighborhood-level missingness. For sensor-level
missingness, we randomly select 5\%, 10\%, or 20\% of the sensors and
mask their entire input histories. For neighborhood-level missingness,
we randomly select a center sensor and mask the input histories of
the center and all sensors within $k$ hops, where $k\in\{1,2,3\}$,
using the PEMS04 adjacency matrix to determine connectivity.
In both settings, masked observations are replaced with zero in the
normalized space, and MAE is evaluated over the full prediction horizon.

As shown in Table~\ref{tab:structured-missing}, STMemoLLM achieves lower
MAE than STD-PLM across all six missingness settings. Under sensor-level
missingness, the relative MAE reductions range from 10.12\% to 13.80\%.
Under neighborhood-level missingness, the reductions range from 11.38\%
to 13.80\%, showing that the performance advantage is maintained when
missing observations are concentrated in connected sensor regions.

\begin{table}[t]
    \centering
    \caption{MAE results under missing settings.}
    \label{tab:structured-missing}
    \resizebox{\linewidth}{!}{
    \begin{tabular}{l|ccc|ccc}
        \toprule
        Dataset & \multicolumn{3}{c|}{Sensor-level missing} & \multicolumn{3}{c}{Neighborhood-level missing} \\
        \cmidrule(lr){1-1} \cmidrule(lr){2-4} \cmidrule(lr){5-7}
        Ratio & 5\% & 10\% & 20\% & 1-hop & 2-hop & 3-hop \\
        \midrule
        STD-PLM & 26.45 & 31.63 & 41.82 & 21.46 & 22.94 & 24.46 \\
        STMemoLLM & \textbf{22.80} & \textbf{27.64} & \textbf{37.59} & \textbf{19.02} & \textbf{19.95} & \textbf{21.08} \\
        Improv. & 13.80\% & 12.59\% & 10.12\% & 11.38\% & 13.04\% & 13.80\% \\
        \bottomrule
    \end{tabular}}
\end{table}

\textbf{Robustness to few-shot settings.}
We evaluate data efficiency by retaining only the first 5\% or 10\%
of the chronologically ordered training split, while keeping the
validation and test splits unchanged. Normalization statistics are
computed using only the retained training data. For each training-data
ratio, STMemoLLM and STD-PLM are trained independently using the same
optimization and early-stopping protocol.

As shown in Table~\ref{tab:few-shot}, STMemoLLM achieves lower MAE than
STD-PLM on all four datasets at both training-data ratios, with relative
MAE reductions ranging from 3.98\% to 15.70\%. It also exhibits smaller
performance degradation relative to full-data training. With 5\% of the
training data, its average relative MAE increase across the four datasets
is 49.4\%, compared with 61.3\% for STD-PLM. With 10\% of the training
data, the corresponding increases are 27.5\% and 35.1\%, respectively.

\begin{table}[t]
    \centering
    \caption{MAE results under few-shot settings.}
    \label{tab:few-shot}
    \resizebox{\linewidth}{!}{
    \begin{tabular}{l|cc|cc|cc|cc}
        \toprule
        Dataset & \multicolumn{2}{c|}{PEMS03} & \multicolumn{2}{c|}{PEMS04} & \multicolumn{2}{c|}{PEMS07} & \multicolumn{2}{c}{PEMS08} \\
        \cmidrule(lr){1-1} \cmidrule(lr){2-3} \cmidrule(lr){4-5} \cmidrule(lr){6-7} \cmidrule(lr){8-9}
        Ratio & 5\% & 10\% & 5\% & 10\% & 5\% & 10\% & 5\% & 10\% \\
        \midrule
        STD-PLM & 22.52 & 17.08 & 27.73 & 25.11 & 32.48 & 26.37 & 22.56 & 19.72 \\
        STMemoLLM & \textbf{20.64} & \textbf{16.33} & \textbf{26.03} & \textbf{22.83} & \textbf{27.38} & \textbf{23.79} & \textbf{21.66} & \textbf{18.90} \\
        Improv. & 8.37\% & 4.42\% & 6.13\% & 9.10\% & 15.70\% & 9.77\% & 3.98\% & 4.15\% \\
        \bottomrule
    \end{tabular}}
\end{table}

\begin{table*}[t]
    \centering
    \caption{MAE results for router ablations.}
    \label{tab:ablation-router}
    \resizebox{0.8 \textwidth}{!}{
    \begin{tabular}{c|c|ccccc}
        \toprule
        Dataset & Full & w/o MCDR & w/o $\boldsymbol{\gamma}_{\text{D}}$ & w/o $\boldsymbol{\gamma}_{\text{C}}$ & w/o $\boldsymbol{\gamma}_{\text{A}}$ & Rand. $\boldsymbol{I}_{\text{F}}$ \\
        \midrule
        PEMS04 & $\mathbf{18.012\pm0.008}$ & $18.096\pm0.005$ & $18.076\pm0.005$ & $18.146\pm0.009$ & $18.052\pm0.004$ & $18.096\pm0.005$ \\
        PEMS08 & $\mathbf{13.170\pm0.007}$ & $13.248\pm0.011$ & $13.244\pm0.005$ & $13.220\pm0.007$ & $13.252\pm0.011$ & $13.242\pm0.004$ \\
        \bottomrule
    \end{tabular}}
\end{table*}

\begin{table*}[t]
    \centering
    \caption{MAE results for preference-bias ablations.}
    \label{tab:ablation-bias}
    \resizebox{0.8 \textwidth}{!}{
    \begin{tabular}{c|c|ccccc}
        \toprule
        Dataset & Full & w/o $\boldsymbol{B}$ & Topo. $\boldsymbol{B}$ & Learn. $\boldsymbol{B}$ & only $\boldsymbol{B}_{\text{pos}}$ & only $\boldsymbol{B}_{\text{neg}}$ \\
        \midrule
        PEMS04 & $\mathbf{18.012\pm0.008}$ & $18.234\pm0.009$ & $18.134\pm0.017$ & $18.064\pm0.019$ & $18.028\pm0.008$ & $18.026\pm0.011$ \\
        PEMS08 & $\mathbf{13.170\pm0.007}$ & $13.292\pm0.004$ & $13.340\pm0.000$ & $13.228\pm0.004$ & $13.242\pm0.013$ & $13.260\pm0.000$ \\
        \bottomrule
    \end{tabular}}
\end{table*}

\begin{table}[t]
    \centering
    \caption{MAE results for loss ablations.}
    \label{tab:ablation-loss}
    \resizebox{\linewidth}{!}{
    \begin{tabular}{c|cccc}
        \toprule
        Dataset & Full & w/o $\mathcal{L}_{\text{nce}}$ & w/o $\mathcal{L}_{\text{dir}}$ & w/o both \\
        \midrule
        PEMS04 & $\mathbf{18.012\pm0.008}$ & $18.064\pm0.011$ & $18.132\pm0.008$ & $18.312\pm0.008$ \\
        PEMS08 & $\mathbf{13.170\pm0.007}$ & $13.248\pm0.008$ & $13.292\pm0.008$ & $13.372\pm0.008$ \\
        \bottomrule
    \end{tabular}}
\end{table}

\subsection{Ablation Study}
We evaluate the contributions of three components of STMemoLLM on PEMS04 and PEMS08: the memory-calibrated router, the node-prototype preference bias, and the training regularizers. All results in
Tables~\ref{tab:ablation-router}-\ref{tab:ablation-loss} are reported as mean MAE $\pm$ sample standard deviation over five random seeds.

\textbf{Memory-calibrated router.}
Table~\ref{tab:ablation-router} compares the full model with variants
that remove the router (w/o MCDR) or replace its selection with random
selection (Rand. $\boldsymbol{I}_{\text{F}}$). We also remove each routing
criterion separately: contextual deviation
($\boldsymbol{\gamma}_{\text{D}}$), compression deficiency
($\boldsymbol{\gamma}_{\text{C}}$), and prototype alignment
($\boldsymbol{\gamma}_{\text{A}}$). All variants yield higher mean MAE
than the full model on both datasets. Among the three criteria, removing
compression deficiency causes the largest increase on PEMS04, whereas
removing prototype alignment has the largest effect on PEMS08. These
results indicate that all three criteria contribute to prediction
accuracy, although their relative importance varies across datasets.

\textbf{Node-prototype preference bias.}
Table~\ref{tab:ablation-bias} examines whether the proposed bias
formulation is needed. We remove $\boldsymbol{B}$, replace it with a
topology-based or learnable node-level bias, and retain only its positive
or negative term. The complete formulation achieves the lowest mean MAE
on both datasets. Removing the bias causes the largest degradation among
these variants on PEMS04, while the topology-based replacement performs
worst on PEMS08. The learnable replacement and both single-term variants
also yield higher mean MAE than the full model. In particular, combining
the positive and negative terms outperforms retaining either term alone,
although the differences are small on PEMS04.

\textbf{Training regularizers.}
Table~\ref{tab:ablation-loss} evaluates the memory contrastive loss
$\mathcal{L}_{\text{nce}}$ and the degree-aware assignment loss
$\mathcal{L}_{\text{dir}}$ by removing each individually and then both
together. Removing $\mathcal{L}_{\text{nce}}$ increases mean MAE by
0.052 and 0.078 on PEMS04 and PEMS08, respectively, while removing
$\mathcal{L}_{\text{dir}}$ produces larger increases of 0.120 and 0.122.
Removing both regularizers leads to the largest degradation, with MAE
increases of 0.300 and 0.202. These results support retaining both
regularizers, with the degree-aware assignment loss showing a larger
individual effect on prediction accuracy.

\subsection{Hyperparameter Study}
\begin{figure}[t]
    \centering
    \includegraphics[width=\columnwidth]{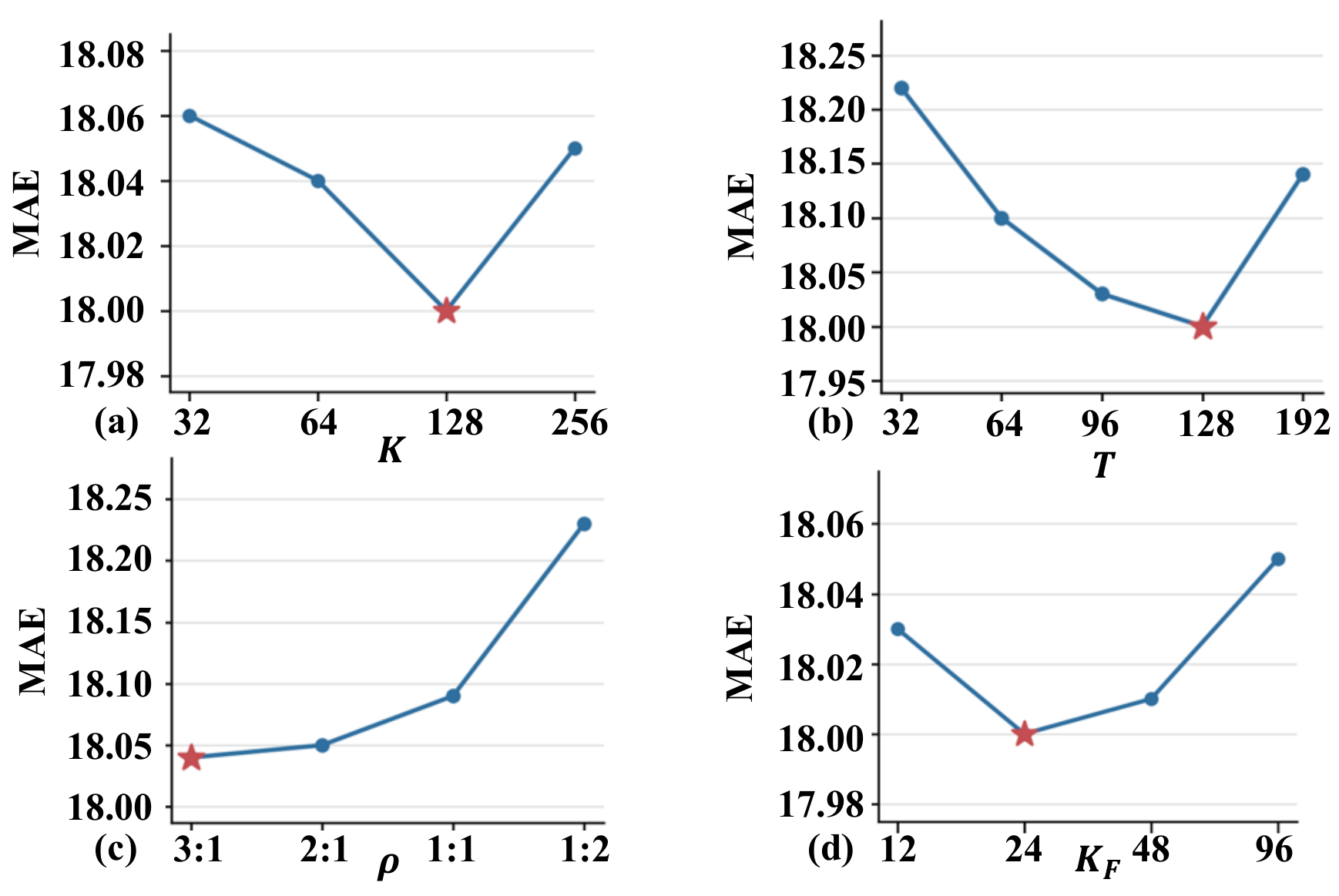}
    \caption{MAE results of the hyperparameter study.}
    \label{fig:hyperparameter-study}
\end{figure}

We evaluate sensitivity to key hyperparameters related to memory size and the trade-off between global compression and local refinement on PEMS04. 
(1) \textbf{Memory prototype number $K$}. As shown in Fig.~\ref{fig:hyperparameter-study}(a), the best value is 128; a smaller $K$ may underfit heterogeneous patterns, while a larger $K$ may fragment memory and harm retrieval stability.
(2) \textbf{Total token numbers $\boldsymbol{T}=N_{\text{sum}}+N_{\text{detail}}$ and summary/detail token split ratio $\rho = N_{\text{sum}}:N_{\text{detail}}$}. As shown in Fig.~\ref{fig:hyperparameter-study}(b) and (c), the best setting is $\boldsymbol{T}=128$ and $\rho=3:1$, balancing global structure and residual dynamics.
(3) \textbf{Routed node number $K_{\text{F}}$}. As shown in Fig.~\ref{fig:hyperparameter-study}(d), the best value is 24; a too small $K_{\text{F}}$ may miss useful details, while a too large value may introduce noise.

\subsection{Efficiency Study}
We compare STMemoLLM with representative pretrained-model-based and classical
graph-based methods in terms of training time, parameter count, peak GPU memory,
and PEMS04 MAE. All models are measured on the same split with batch size 64 on
an RTX 5090 using the same timing and memory procedure. Trainable parameter counts exclude frozen backbone weights and therefore do not quantify total model-storage requirements.
The comparison exposes an accuracy-resource trade-off. STGCN and Graph WaveNet are substantially faster
and use less memory than STMemoLLM, while DCRNN has a similar epoch time and
MegaCRN requires a longer training time. STMemoLLM obtains the lowest MAE in this
comparison, at the cost of a larger parameter count and higher peak memory than
the lightweight graph models. It is therefore most suitable when forecasting
accuracy is prioritized over resource minimization. 

\begin{table}[t]
    \centering
    \caption{MAE results of the efficiency study.}
    \label{tab:efficiency}
    \resizebox{\linewidth}{!}{
    \begin{tabular}{lcccc}
        \toprule
        Method & Time (s/epoch) & Params (M) & Memory (GB) & MAE \\
        \midrule
        DCRNN & 20.72 & 0.37 & 4.93 & 21.22 \\
        STGCN & 1.83 & 0.39 & 0.80 & 21.16 \\
        Graph WaveNet & 9.68 & 0.31 & 2.19 & 24.89 \\
        MegaCRN & 67.08 & 0.39 & 4.91 & 18.78 \\
        \midrule
        STLLM & 11.28 & 42.5 & 5.10 & 19.00 \\
        TimeCMA & 75.40 & 29.4 & 21.63 & 21.79 \\
        STLLM+ & 42.11 & 3.2 & 12.01 & 18.58 \\
        STD-PLM & 20.32 & 3.1 & 11.80 & 18.20 \\
        \midrule
        STMemoLLM & 22.39 & 4.60 & 11.37 & \textbf{18.01} \\
        \bottomrule
    \end{tabular}}
\end{table}

\subsection{Analysis of Summary Representations and Selective Detail Routing}
STMemoLLM uses prototype-guided summary tokens to represent recurring traffic
patterns and routed detail tokens to retain residual dynamics. We examine
these roles on PEMS04 through recurring-context organization, controlled
perturbations, and forecasting performance in naturally occurring tail regimes.

\begin{figure}[!t]
    \centering
    \includegraphics[width=\columnwidth]{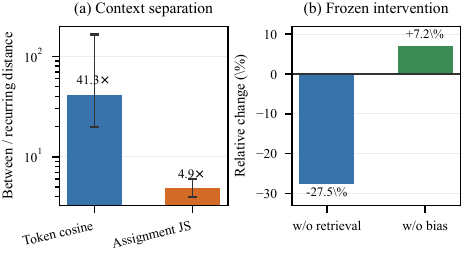}
    \caption{Recurring-context organization on PEMS04. (a) Ratios of six-hour-shifted to weekly recurring pair distances; error bars show 95\% anchor-day bootstrap intervals. (b) Relative changes in the summary separation gap after removing retrieval and in recurring-context assignment JS after removing the preference bias.}
    \label{fig:recurring-context}
\end{figure}

\textbf{Summary tokens organize recurring contexts.}
We construct 512 held-out triples across 16 anchor days. Each triple contains
an anchor window, a recurring window at the same time-of-day slot seven days
later, and a comparison window shifted by six hours on the anchor day.
The shifted-to-recurring distance ratios are $41.28\times$ for summary-token
cosine distance and $4.88\times$ for assignment JS divergence, with 95\%
anchor-day bootstrap intervals of $[19.91,165.30]$ and $[3.95,5.99]$,
respectively (Fig.~\ref{fig:recurring-context}(a)).
With the checkpoint, inputs, and timestamps fixed, removing retrieval reduces
the summary separation gap by $27.5\%$ ($0.898 \to 0.651$), while removing
the node--prototype preference bias increases recurring-context assignment JS
by $7.2\%$ ($0.000430 \to 0.000461$; Fig.~\ref{fig:recurring-context}(b)).
These results support the roles of retrieval in separating traffic contexts
and the preference bias in stabilizing assignments across recurring contexts.

\begin{figure}[!t]
    \centering
    \includegraphics[width=\columnwidth]{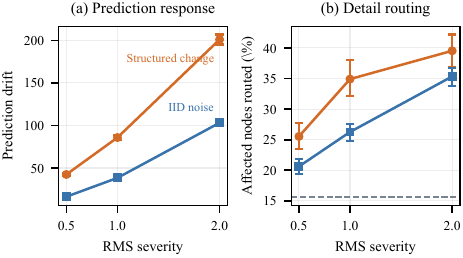}
    \caption{Responses to energy-matched structured changes and i.i.d. noise on PEMS04. (a) Prediction drift on affected nodes. (b) Fraction of affected nodes routed to the detail path; the dashed line marks random routing. Error bars show 95\% paired-case bootstrap intervals.}
    \label{fig:structured-routing}
\end{figure}

\textbf{Detail routing responds to structured deviations.}
Using a periodic baseline estimated from the training split, we construct
256 paired inputs with either a spatially coherent, temporally persistent
change or i.i.d. Gaussian noise. Both perturbations affect the same connected
30-sensor region over the final four history steps and have identical support
and standardized $L_2$ energy.
Structured changes produce greater prediction drift on affected nodes and
higher affected-node routing rates across all three severities
(Fig.~\ref{fig:structured-routing}). At RMS severity 1, the prediction drift
is $2.24\times$ that under noise (95\% CI $[2.15,2.32]$), and the routing
rates are $34.9\%$ and $26.3\%$, respectively. The paired routing difference
is $8.65$ percentage points (95\% CI $[6.20,11.07]$).
These results indicate that detail allocation is sensitive to spatial and
temporal organization beyond perturbation energy alone.

\textbf{Selective detail routing improves tail-regime forecasts.}
We define rare and extreme node--windows using sensor-specific training P95
and P99 thresholds, respectively, for traffic amplitude or temporal change.
Low alignment (LA) and high deviation (HD) are defined using sensor-specific
validation P20 and P80 thresholds for prototype alignment and contextual
deviation, respectively. To isolate the detail contribution, we set it to zero
at inference while keeping the remaining computation fixed, and measure MAE
reductions relative to this summary-only output.

\begin{figure}[t]
    \centering
    \includegraphics[width=\columnwidth]{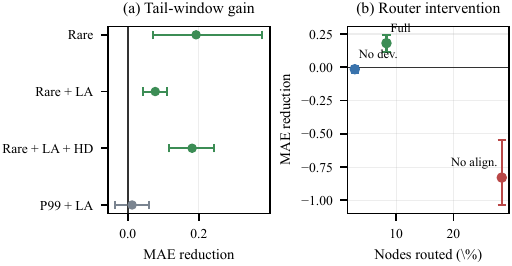}
    \caption{Tail-regime evaluation on PEMS04. (a) MAE reductions relative to the summary-only output; error bars show 95\% confidence intervals from 1,000 paired day-block bootstrap resamples. LA and HD denote low alignment and high deviation. (b) Routing rates and MAE reductions under router interventions on rare LA+HD node--windows.}
    \label{fig:tail-windows}
\end{figure}

The detail path reduces MAE by $0.192$ on rare node--windows and by $0.181$
on the rare LA+HD subset, compared with $0.077$ on all rare LA node--windows
(Fig.~\ref{fig:tail-windows}(a)). For the P99 LA group, however, the gain is
only $0.012$ and its confidence interval crosses zero, leaving the benefit
for this extreme group inconclusive.
On the rare LA+HD subset, removing contextual deviation lowers the routing
rate from $8.3\%$ to $2.8\%$ and changes the MAE reduction from $0.181$ to
$-0.015$. Removing alignment raises the routing rate to $28.4\%$ and changes
the reduction to $-0.829$ (Fig.~\ref{fig:tail-windows}(b)).
These interventions suggest complementary roles: contextual deviation helps
identify useful local dynamics, while alignment limits unhelpful detail
allocation under weak prototype support.

\begin{figure}[t]
    \centering
    \includegraphics[width=\columnwidth]{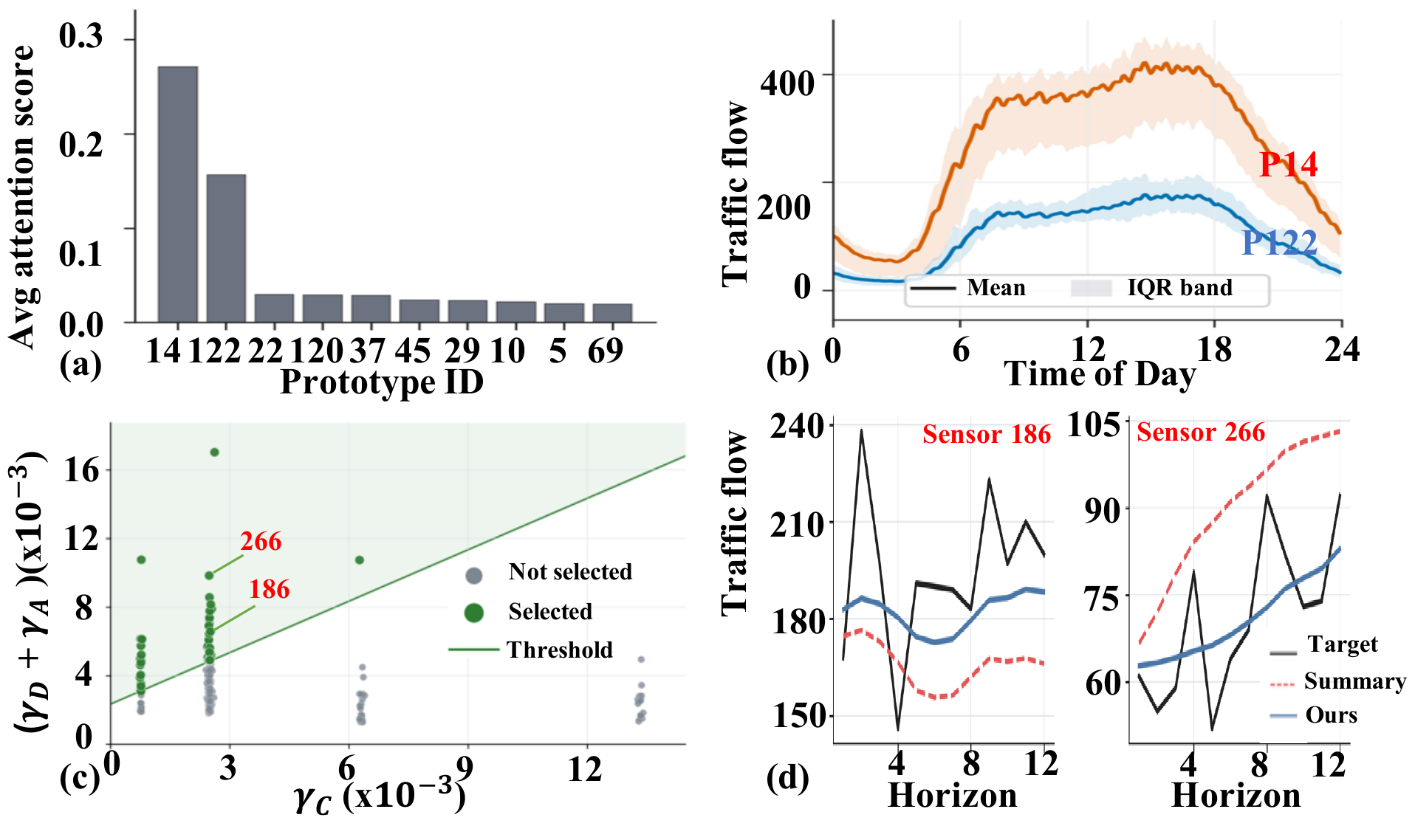}
    \caption{Results of the case study.}
    \label{fig:memory-interpretability}
\end{figure}

\subsection{Case Study}
We conduct the case study on PEMS04. Fig.~\ref{fig:memory-interpretability}(a) visualizes the top-10 retrieved prototypes based on $\boldsymbol{a}$ in Eq.~\eqref{eq:memattn}. Retrieval concentrates on a compact subset without collapsing to a single prototype, and the top-10 prototypes account for $0.61$ of total attention mass, indicating effective yet non-degenerate memory usage.
Based on the frequent retrieval of prototypes \#14 and \#122 in (a), Fig.~\ref{fig:memory-interpretability}(b) further analyzes their semantics by aggregating associated sensors. The mean is the attention-weighted average, and the band denotes the IQR (25th–75th percentile). \#14 corresponds to higher flow with stronger temporal variation, while \#122 captures lower flow with weaker dynamics, showing that the memory encodes diverse and interpretable traffic states.
Fig.~\ref{fig:memory-interpretability}(c) illustrates the tri-criteria filtering process for identifying detail tokens, where the x-axis denotes coverage $\boldsymbol{\gamma}_{\text{C}}$ and the y-axis jointly represents prototype alignment $\boldsymbol{\gamma}_{\text{A}}$ and contextual deviation $\boldsymbol{\gamma}_{\text{D}}$. Specifically, nodes above the diagonal selection threshold (shown in green) are preserved, such as Nodes 266 and 186. The corresponding predictive performance for these two nodes is visualized in Fig.~\ref{fig:memory-interpretability}(d). A comparison between the use of summary tokens alone (red dashed lines) and the integration of summary and detail tokens (blue solid lines) reveals that the latter more closely tracks the ground-truth. These results from Fig.~\ref{fig:memory-interpretability}(c) and (d) collectively validate the effectiveness of incorporating detail tokens for accurate forecasting.

\section{Conclusions and Future Work}
In this work, we proposed STMemoLLM, a spatial–temporal memory-augmented large language model for robust traffic forecasting. Leveraging memory prototypes with memory-enhanced tokenization and memory-calibrated routing, STMemoLLM compresses correlated sensor observations into compact semantic representations while selectively incorporating reliable residual dynamics. Extensive experiments on eight real-world datasets demonstrate that STMemoLLM achieves the SOTA forecasting performance on most evaluated metrics.
Additional evaluations show improved robustness
under point masking, Gaussian noise, structured missingness, and few-shot settings. Representation and routing analyses further support the
complementary roles of recurring-context summaries and selectively
retained local dynamics.

Since STMemoLLM requires more computational and memory resources than lightweight graph models, future work will focus on reducing training and memory costs for large-scale traffic networks. In addition, the benefits of STMemoLLM in extreme regimes with weak prototype support remain inconclusive. Future work will also improve the modeling of rare dynamics beyond the existing prototype space.

\clearpage

\section*{AI-Generated Content Acknowledgement}
During manuscript preparation, ChatGPT (OpenAI) was used as a language-editing aid for limited English expression and grammar refinement. The research conception, methodology, implementation, evaluation, interpretation of results, and conclusions were developed and verified by the authors.

\bibliographystyle{IEEEtran}
\bibliography{references}

\end{document}